\documentclass{bmvc2k}

\usepackage{amsmath}
\usepackage{amssymb}

\usepackage{hyperref}
\usepackage{url}
\usepackage{graphicx}
\usepackage{subfigure}
\usepackage{multirow}
\usepackage{xcolor}
\usepackage{booktabs} 

\usepackage{adjustbox}

\usepackage{floatrow}

\newcommand{\ie}{\textit{i}.\textit{e}., }

\title{TransFusion: Cross-view Fusion with Transformer for 3D Human Pose Estimation}

\addauthor{Haoyu Ma}{haoyum3@uci.edu}{1}
\addauthor{Liangjian Chen}{clj@fb.com}{2}
\addauthor{Deying Kong}{deyingk@uci.edu}{1}
\addauthor{Zhe Wang}{zwang15@uci.edu}{1}
\addauthor{Xingwei Liu}{xingweil@uci.edu}{1}
\addauthor{Hao Tang}{htang6@uci.edu}{1}
\addauthor{Xiangyi Yan}{xiangyy4@uci.edu}{1}
\addauthor{Yusheng Xie}{yushx@amazon.com}{3}
\addauthor{Shih-Yao Lin}{mike.lin@ieee.org}{4}
\addauthor{Xiaohui Xie}{xhx@uci.edu}{1}

\addinstitution{
 University of California, Irvine
}
\addinstitution{
 Facebook \\
 (work done outside of Facebook)
}
\addinstitution{
 Amazon \\ (work done outside of Amazon)
}
\addinstitution{
 R\&D Center~US~Laboratory,\\
 Sony~Corporation~of~America \\
 (work done outside of Sony)
}

\runninghead{ H. Ma et al.}{TransFusion: Cross-view fusion with Transformer}

\begin{document}

\maketitle

\begin{abstract}
Estimating the 2D human poses in each view is typically the first step in calibrated multi-view 3D pose estimation. But the performance of 2D pose detectors suffers from challenging situations such as occlusions and oblique viewing angles. To address these challenges, previous works derive point-to-point correspondences between different views from epipolar geometry and utilize the correspondences to merge prediction heatmaps or feature representations. Instead of post-prediction merge/calibration, here we introduce a transformer framework for multi-view 3D pose estimation, aiming at directly improving individual 2D predictors by integrating information from different views. Inspired by previous multi-modal transformers, we design a unified transformer architecture, named TransFusion, to fuse cues from both current views and neighboring views. Moreover, we propose the concept of {\em epipolar field} to encode 3D positional information into the transformer model.  The 3D position encoding guided by epipolar field provides an efficient way of encoding correspondences between pixels of different views. Experiments on Human 3.6M and Ski-Pose show that our method is more efficient and has consistent improvements compared to other fusion methods. Specifically, we achieve 25.8 mm MPJPE on Human 3.6M with only 5M parameters on 256 $\times$ 256 resolution. 
Source code and trained model can be found at \href{https://github.com/HowieMa/TransFusion-Pose}{https://github.com/HowieMa/TransFusion-Pose}. 
\end{abstract}

\vspace{-1.0em}
\section{Introduction}
\vspace{-0.7em}
Estimating the 3D locations of human joints is a critical task for many AI applications such as augmented reality, virtual reality and medical diagnosis \cite{chen2021pd}. The estimation is often carried out in two common settings: One is estimating the 3D pose from monocular images \cite{mehta2017vnect, zhang2019end, zimmermann2019freihand, ge20193d, wang2020predicting, chen2020dggan, chen2018generating, tome2017lifting}, and the other is estimating 3D poses from multiple cameras \cite{simon2017hand, wang2019geometric, chen2021mvhm, qiu2019cross, he2020epipolar}. The former is challenging due to the ambiguity of depth estimation with only one view. The latter setting, the focus of this paper, usually obtains better 3D pose estimation performance since the multi-view settings can help resolve depth ambiguity. Most multi-view works follow a two-step pipeline that firstly estimates 2D poses in each view and then recovers 3D pose from them. However, it is still difficult to solve challenging cases such as occlusions in the first step, and the estimated 3D poses are often inaccurate as it depends on the results from the first step. 

\begin{figure}[!t]
\floatbox[{\capbeside\thisfloatsetup{capbesideposition={left,top},capbesidewidth=0.5\textwidth}}]{figure}[\FBwidth]
{\caption{ \footnotesize{Comparison of epipolar line and attention module. Given the query pixel (cyan dot) in view 1 (current). The attention map on the view 2 (reference) indicates that the prediction relies on the image clues provided by the area of right shank, not just the corresponded right ankle. While previous methods based on epipolar line (yellow line) cannot capture this information.   }}\label{fig:test}}
{\includegraphics[width=0.4\textwidth]{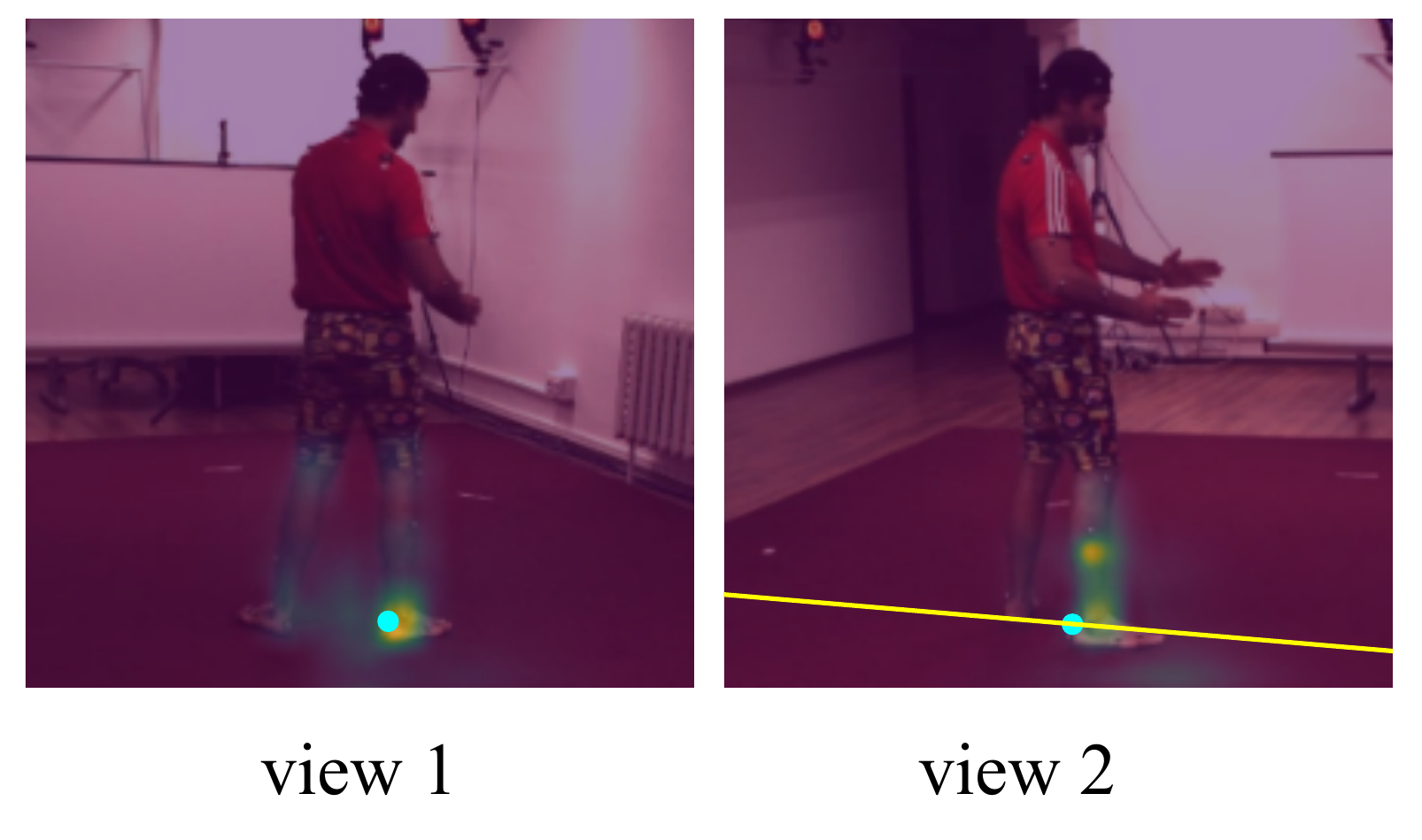}}
\label{fig:att_map}
\end{figure}

Researchers have sought to introduce the 3D information in the first step to improve the 2D pose detector, because the challenging cases in one view are potentially easier to solve in other views. Specifically, they usually fuse the features of the neighboring view (reference view) with epipolar constraints \cite{xie2020metafuse, zhang2021adafuse, he2020epipolar}. 
Although interpretable, fusing along the epipolar line only does not fully utilize the semantic information of the reference view as the information off the epipolar line is discarded. 
For example, it is difficult to associate the ankle with the leg from the epipolar line in the reference view of Figure \ref{fig:att_map}, which could be an important cue as part of the structure information for pose estimation. 
On the other hand, fusing all locations of other views can address this drawback. In this paper, we propose the  \textit{Epipolar Field}, a more general form of the epipolar line. It assigns probabilities to all locations of the reference view and still keep the knowledge of epipolar constraints.

Recently, attention mechanisms and the transformers \cite{vaswani2017attention} achieve great progress in computer vision areas \cite{wang2018non, dosovitskiy2020image, carion2020end, zhu2020deformable, zheng2020rethinking, lin2020end, tang2021spatial, wang2022sscap}.
The self-attention module \cite{vaswani2017attention} can capture long range dependencies and correspondences, which is difficult for the convolutional layer. Although promising, there are only a few works \cite{lin2020end} that apply it to the 3D pose estimation tasks. To the best of our knowledge, none of the previous works have exploited the transformer architectures in the multi-view 3D pose estimation setting. Inspired by previous multi-modal transformers \cite{su2019vl, tan2019lxmert, kim2021vilt}, we propose the \textit{TransFusion}, a lightweight framework that can utilize all pixels from both the current view itself and reference view simultaneously. As an example in Figure \ref{fig:att_map}, the attention layer actually relies on the whole leg to infer the location of the ankle. Moreover, we add the 3D geometry positional encoding based on the epipolar field to help the transformer explicitly capture the correspondence.

Our main contributions are summarized as follows:
\vspace{-0.5em}
\begin{itemize}
    \item We are the first to apply the transformer architecture to multi-view 3D human pose estimation. We propose the TransFusion, a unified architecture to fuse cues from multiple views. 
    
    \vspace{-0.5em}
    \item We propose the \textit{epipolar field}, a novel and more general form of epipolar line. It readily integrates with the transformer through our proposed geometry positional encoding to encode the 3D relationships among different views.  
    
    \vspace{-0.5em}
    \item Extensive experiments are conducted to demonstrate that our TransFusion  outperforms previous fusion methods on both Human 3.6M and SkiPose datasets, but requires substantially fewer parameters.
    \vspace{-0.5em}
\end{itemize}

\section{Related Work}
\vspace{-0.7em}
\subsection{Multi-view 3D Pose Estimation}
\vspace{-0.6em}
Multi-view 3D pose estimation usually follows a two-step process: (1) localize 2D joints with a 2D pose estimator on each view, and (2) lift the 2D joints from multi-view images to the 3D position via triangulation. 
To improve the performance of 2D pose detector, researchers typically resort to sophisticated architectures to capture both low-level and high-level representations  \cite{wei2016convolutional, chen2018cascaded, newell2016stacked, xiao2018simple, sun2019deep} or use the structural information to model the spatial constraints \cite{tompson2014joint, kong2019adaptive, kong2020rotation, kong2020sia, chen2020nonparametric}. However, the occlusion cases are still challenging, as monocular images do not provide evidence for occlusion joints localization.

An alternative approach, more explainable, is to make the 2D pose detector 3D-aware, \ie fusing the 2D feature heatmaps \cite{qiu2019cross, zhang2021adafuse, xie2020metafuse, chen2021mvhm, he2020epipolar} from different views. 
Specifically, the Cross-view Fusion \cite{qiu2019cross} directly learns a fixed attention weight to fuse all pairs of pixels given a pair of views. However, the learnable weight requires the multi-camera setup unchanged during the inference time, and the number of parameters is quadratic to the resolution of input images. 
The epipolar transformer \cite{he2020epipolar} applies the non-local module \cite{wang2018non} to obtain the weights and only fuse pixels along the epipolar line in other views. Thus it is easy to learn and flexible to use. 
However, sampling along the epipolar line discards off-epipolar line information and thus obtains limited information from the reference views.
In the second step, researchers use graphical model with the structure of human \cite{qiu2019cross} to improve the quality of triangulation or directly learn 3D pose via differentiable triangulation \cite{iskakov2019learnable}. 
Our work still focuses on enhancing 2D pose by fully integrating information from different views.

\vspace{-1.0em}
\subsection{Transformer}
\vspace{-0.5em}
\paragraph{Vision Transformer} Recently, several studies demonstrated that the transformer architectures \cite{vaswani2017attention} plays a significant role in a wide range of computer vision tasks, such as image classification \cite{dosovitskiy2020image, touvron2020training, chen2021crossvit}, object detection \cite{carion2020end, zhu2020deformable}, and semantic segmentation \cite{zheng2020rethinking, wang2020end, yan2022after}. 
Recently, some studies also explored applying the transformer on human pose estimation tasks \cite{yang2020transpose,li2021pose, mao2021tfpose, lin2020end, zheng20213d}. 
More specifically, for 2D pose estimation, TransPose \cite{yang2020transpose} aims to explain the spatial dependencies of the predicted keypoints with transformers, PRTR \cite{li2021pose} and TF-Pose \cite{mao2021tfpose} attempt to directly regress the joint coordinates by transformer decoders. 
While for the 3D pose estimation, METRO \cite{lin2020end} firstly applies transformer to reconstruct 3D human pose and mesh from a single image, and PoseFormer \cite{zheng20213d} builds a spatial-temporal transformers with the input of 2D joint sequences for 3D pose estimation in videos. 
However, previous works have hardly exploited the transformer architectures on the multi-view 3D pose estimation setting, which is however an important task in the pose estimation area.

\vspace{-1.0em}
\paragraph{Multi-modal Transformer}
Transformers with multi-modality inputs such as images and texts have also been fully exploited \cite{kim2021vilt, tan2019lxmert, su2019vl, li2020unicoder, zhuge2021kaleido, you2021mrd, you2021self}. In general, these methods directly concatenate the embeddings from two sources together \cite{kim2021vilt} and make the transformer itself to learn the correspondence between two modalities from  millions of image-text paris \cite{sharma2018conceptual}. 
Thus, these methods are quite expensive and inefficient, and difficult to apply on limited datasets. Our method, however, directly provides the correspondence between two inputs and makes the transformer explicitly learn their relationships.

\vspace{-1.0 em}
\section{Methods}
\vspace{-0.8 em}
\subsection{Overview}
\begin{figure}[!t]
    \centering
    \includegraphics[width=0.9\textwidth]{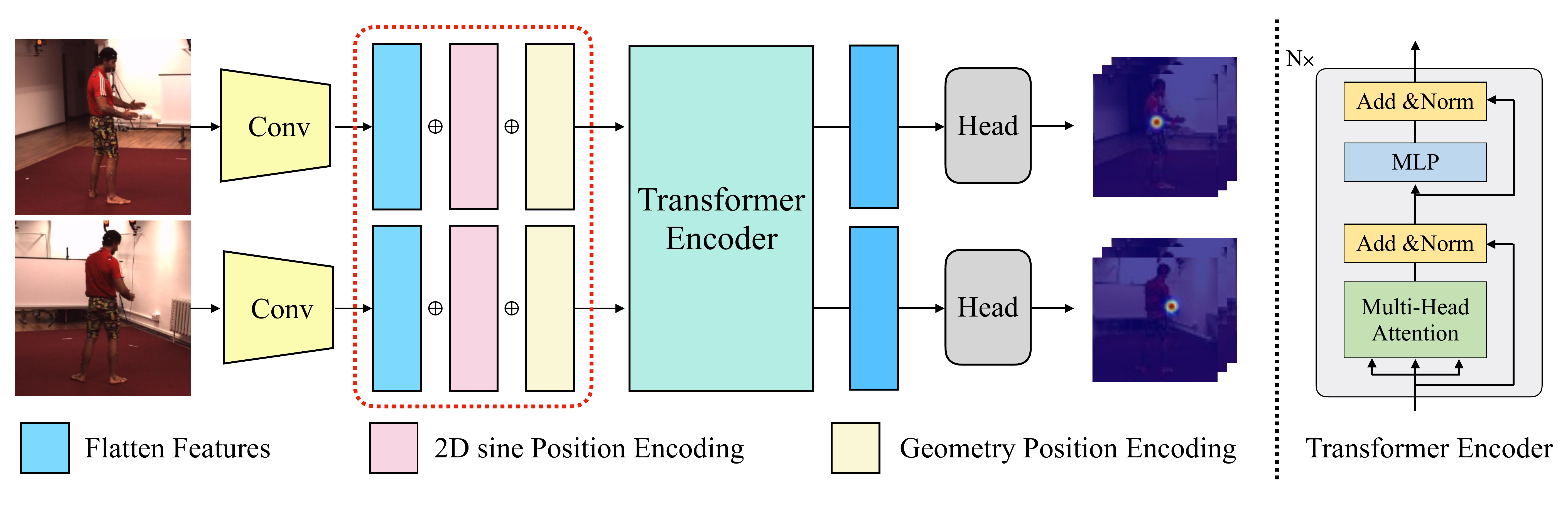}
    \vspace{-1 em}
    \caption{\footnotesize{Overview of TransFusion. }}
    \label{fig:network}
\end{figure}

\vspace{-0.8em}
Figure \ref{fig:network} is an overview of TransFusion. 
It takes two images from different views as input, and predicts the heatmaps of joints in each view. The framework consists of three modules: a CNN backbone to extract low-level features; a transformer encoder to capture both correspondence between two views and long-range spatial correlations within single view images; a head to predict the heatmaps of joints. 
Specifically, given images $ \mathbf{I}_i \in \mathbb{R} ^{3 \times H_I \times W_I}$ in each view, where $i\in\{1, 2\}$ denotes view $1$ and view $2$, the backbone $\mathcal{F}(\cdot)$ firstly produces the low-level features $ \mathbf{X}_i = \mathcal{F}(\mathbf{I}_i) \in \mathbb{R}^{d \times H \times W}$ of each image. Here $d$ is the number of channels.  $H$ and $W$ are the height and width of the feature map, respectively. The feature $ \mathbf{X}_i$ is flattened into a sequence vector $\mathbf{X}_i^{'} \in \mathbb{R}^{L \times d}$, where $L= H \times W$. 
Both 2D sine positional encoding $\mathbf{E_{2D}}$ and 3D geometry positional encoding $\mathbf{E_{G}}_i$ are added onto $\mathbf{X}_i^{'}$ to make the transformer aware of position information. $ \mathbf{X}_1^{'}$ and $\mathbf{X}_2^{'}$ are concatenated together to build a uniform embedding $\mathbf{X} = [\mathbf{X}_1^{'} + \mathbf{E_{2D}} + \mathbf{E_{G}}_1 , \mathbf{X}_2^{'} + \mathbf{E_{2D}} + \mathbf{E_{G}}_2] \in \mathbb{R}^{2L \times d}$. The embedding $\mathbf{X}$ then enters the standard transformer encoder $\mathcal{E}(\cdot)$. 
Finally, the output of the transformers $\mathbf{\tilde{X}}=\mathcal{E}(\mathbf{X})$ are split into $\tilde{\mathbf{X}_1}$ and $\tilde{\mathbf{X}_2}$, which is embedding of each view, and a prediction head $\mathcal{H}(\cdot)$ takes $\tilde{\mathbf{X}_i}$ and predicts the joint heatmaps $\bar{\mathbf{H}}_i = \mathcal{H}( \tilde{ \mathbf{X}_i} ) \in \mathbb{R}^{k \times H_h \times W_h}$ for each view, where $k$ is the number of joints.

\vspace{-1.0em}
\subsection{TransFusion}
\vspace{-0.5em}
\paragraph{Transformer Encoders}
The transform encoder $\mathcal{E(\cdot)}$ consists of several layers of multi-head self-attention. 
Let $l = 2L$ for short, given the input sequence $\mathbf{X} \in \mathbb{R}^{l \times d}$, the self-attention layer first uses linear projections to obtain a set of queries ($\mathbf{Q} \in \mathbb{R}^{l \times d} $), keys ($\mathbf{K} \in \mathbb{R}^{l \times d} $) and values ($ \mathbf{V} \in \mathbb{R}^{l \times d}$) from $\mathbf{X}$. The three linear projections are parameterized by three learnable matrices $\mathbf{W}_q$,  $\mathbf{W}_k$, $\mathbf{W}_v \in \mathbb{R}^{d\times d}$.
Following \cite{carion2020end}, the position encoding $\mathbf{E}$ is added into the input $\mathbf{X}$ for computing the query and key. 
The scaled dot-product attention \cite{vaswani2017attention} between $\mathbf{Q}$ and $\mathbf{K}$ is adopted to compute the attention weights, and aggregate the values: 
\begin{equation}
    \mathbf{A}=\operatorname{softmax}\left(\frac{\mathbf{Q K}^{T}}{\sqrt{d}}\right) \mathbf{V}
\end{equation}
Finally, a non-linear transformation (\ie multi layer perceptron, and the skip connection) is applied on $\mathbf{A}$ to calculate the output $\tilde{\mathbf{X}}$. 
As $\mathbf{X}$ is low-level features of all views, given one query pixel on the feature map, it can attend cues from the its own view and other views simultaneously through the entire network. 

\vspace{-0.8em}
\paragraph{Positional encoding}
The attention layer would degenerate into a permutation-equivariant architecture without any position information. Thus, the positional encoding is necessary to make the transformer aware of position and order of input sequence.  
For each individual view, we follow the 2D sine positional encoding in the original transformers \cite{dosovitskiy2020image, yang2020transpose}, and we denote it as $\mathbf{E_{2D}}$. However, it only encodes position information from its own view, while the position information in the 3D space and that from the reference views cannot be encoded. Thus, another positional encoding $\mathbf{E_{G}}_i$ (See Section \ref{sec:gpe}) is required to encode the 3D location information of each view $i$ in the 3D space. 


\vspace{-0.5em}
\subsection{Geometry Position Encoding (GPE)}
\begin{figure}[!ht]
    \centering
    \includegraphics[width=0.4\textwidth]{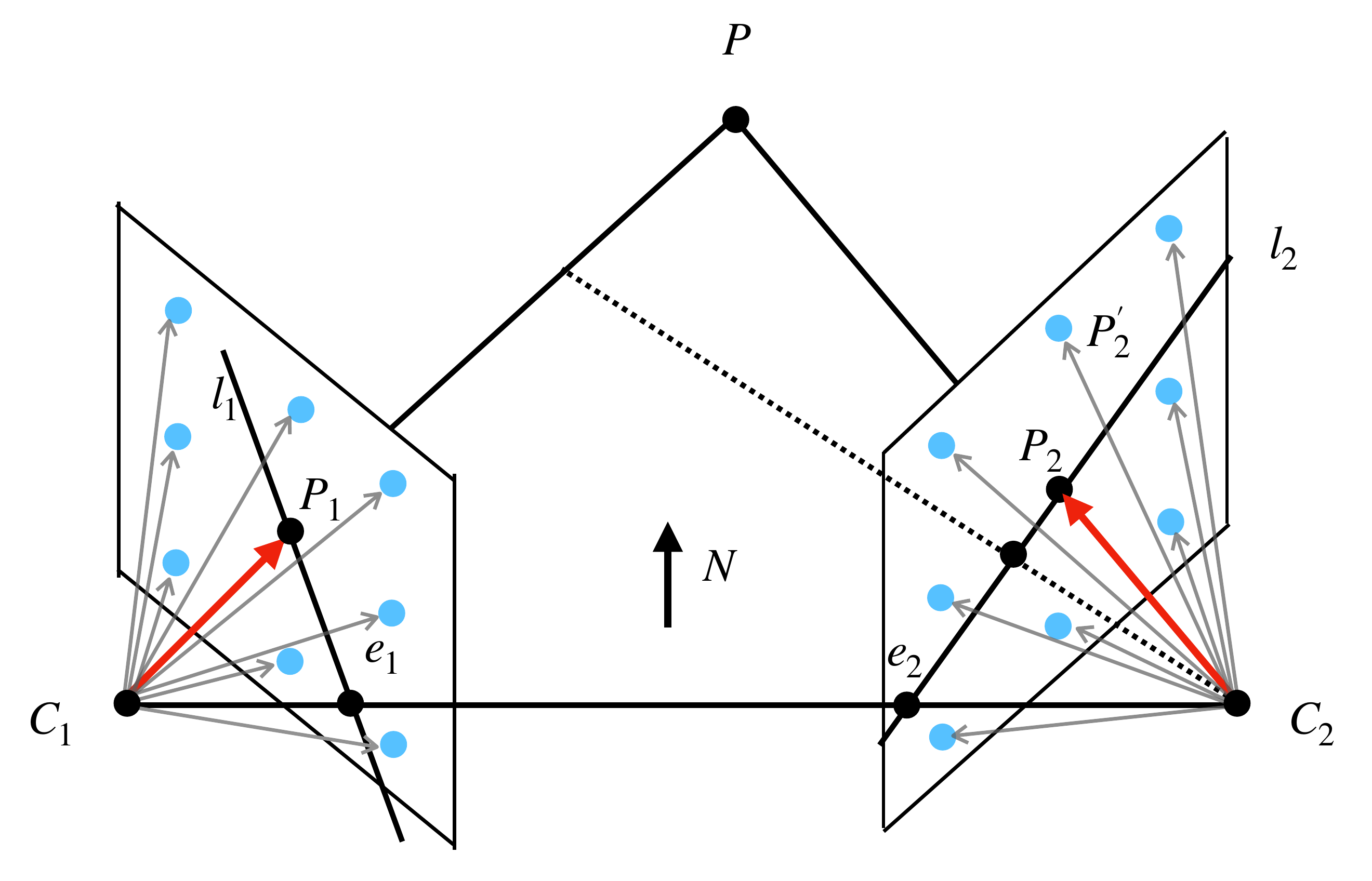}
    \vspace{-1 em}
    \caption{ \footnotesize{Illustration of Geometry Positional Encoding. For each pixel, its geometry positional encoding is calculated from the unit vector $\mathbf{\hat{u}_i^n} $ of ray ${\mathbf{C}_i \mathbf{P}_i}$ in the world coordinate $\mathbf{O}_{\text{world}}$.}}  
    \label{fig:epipolar}
\end{figure}

\vspace{-0.5em}
\label{sec:gpe}
To make the transformers 3D-aware, we introduce 3D camera information \cite{andrew2001multiple, zhao2021camera} into the positional encoding and propose the \textit{Geometry Positional Encoding}. 
Denote the world coordinate system as $\mathbf{O}_{\text{world}}$. The 3D location of view $i$ 's camera center in $\mathbf{O}_{\text{world}}$ is denoted as $\mathbf{C}_i$, and the 3D location of the $n$-th pixel ($n\in \{1,2,..., L\}$) of view $i$ in $\mathbf{O}_{\text{world}}$ is denoted as  $\mathbf{P}_i^n$.  $\mathbf{P}_i^n$ can be derived from the camera parameters of view $i$.  
As shown in Figure \ref{fig:epipolar}, the ray $ \overrightarrow{\mathbf{C}_i \mathbf{P}_i^n}$  (gray line with arrow) indicates the direction of the pixel $p_i$ in the world. 
The unit vector $ \widehat{\overrightarrow{\mathbf{C}_i \mathbf{P}_i^n} } $ is its direction vector, and can encode the relative 3D location of each pixel. 
Thus, we design GPE based on this unit vector, and we add one linear transformation to make it fit the input dimension $d$. The 3D geometry positional encoding for the $n$-th pixel in view $i$ is defined as:   
\begin{equation}
\label{eq:3d_PE}
    \mathbf{E_G}_{i}^{n} = \mathbf{W}_e  \widehat{\overrightarrow{\mathbf{C}_i \mathbf{P}_i^n} }  \in \mathbb{R}^d
\end{equation}
Where $\mathbf{W}_e \in \mathbb{R}^{d \times 3}$ is a learnable transformation matrix. With $\mathbf{E_G}_{i}$, the transformer can be aware of the 3D location of each view.

\vspace{-0.5em}

\vspace{-1.0em}
\subsection{Epipolar Field} 
\begin{figure}[!ht]
    \centering
    \includegraphics[width=0.95\textwidth]{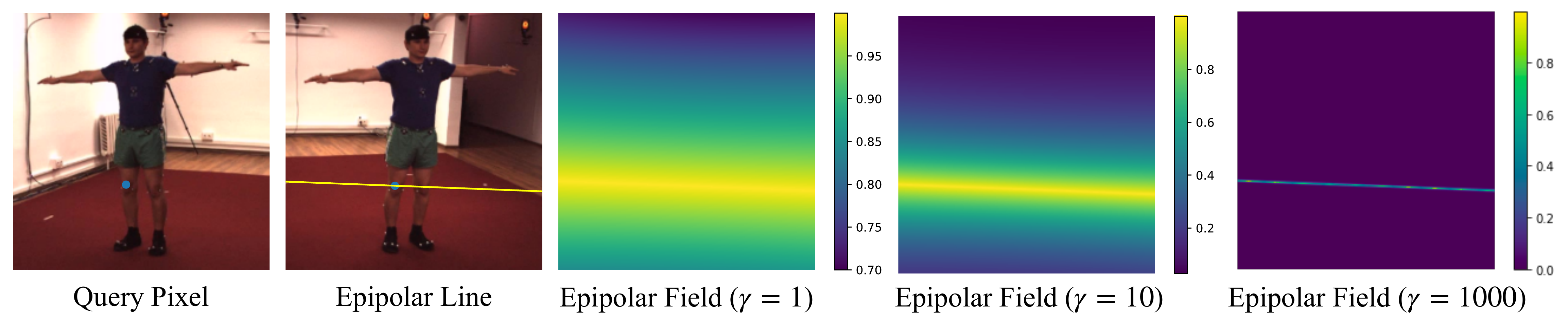}
    \vspace{-1 em}
    \caption{ \footnotesize{Illustration of Epipolar Field. The epipolar field can reflect the distance from the epipolar line to the off-line pixels. By adjusting the soft factor $\gamma$, it can also reduce to the standard epipolar line.}  }
    \label{fig:epipolar_example}
\end{figure}

\vspace{-0.5em}
Although GPE impose the 3D space information into transformers, it does not explicitly encode the relationship between two views. 
As a result, given a pixel in the current view, it is still difficult to attend the corresponding regions when performing global attention between features of two views.  
We further impose the \textit{Epipolar Constraints} \cite{andrew2001multiple} into GPE: Given one pixel $p_1$ in view $1$, its correspondence pixel $p_2$ in view $2$ must be on the epipolar line $l_2$  (Figure \ref{fig:epipolar}). 
However, the epipolar line does not model the relationship with pixels outside $l_2$. Instead, pixels close to the line and pixels away from the line should be treated differently. Thus, we propose the \textit{Epipolar Field} to model the relationship among all pixels in the reference view. In detail, given $\mathbf{P}_1^n$, the 3D location of $n$-th pixel $p_1^n$ in view 1, we calculate  the normal vector $\mathbf{N}_{\mathbf{P}_1^n\mathbf{C}_1\mathbf{C}_2}$ of plane $\mathbf{P}_1^n\mathbf{C}_1\mathbf{C}_2$ by: 
\begin{equation}
    \mathbf{N}_{\mathbf{P}_1^n\mathbf{C}_1\mathbf{C}_2} = \widehat{\overrightarrow{\mathbf{C}_1 \mathbf{C}_2}}  \times  \widehat{\overrightarrow{\mathbf{C}_1 \mathbf{P}_1^n}  }
\end{equation}
Given $\mathbf{P}_2^m$, the 3D location of $m$-th pixel $p_2^m$ in view $2$, we use the angle $\theta$ between the normal vector $\mathbf{N}_{\mathbf{P}_1^n\mathbf{C}_1\mathbf{C}_2}$ and ray  $\overrightarrow{\mathbf{C}_2 \mathbf{P}_2^m}$ to model the relationship between $p_1^n$ and $p_2^m$, and use the cosine of $\theta$ to calculate the correspondence score: 
\begin{equation}
    \mathbf{S}(p_1^n, p_2^m) = 1 - \mid \cos \theta \mid = 1 -  \mid \mathbf{N}_{\mathbf{P}_1^n\mathbf{C}_1\mathbf{C}_2}  \cdot   \widehat{\overrightarrow{\mathbf{C}_2 \mathbf{P}_2^m}} \mid
    \label{eq:score}
\end{equation}
The absolute $\mid \cdot \mid$ is added to limit the score in $[0, 1]$. 
With Eq. \ref{eq:score}, if $p_2^m$ falls in the epipolar line $l_2$, the score $\mathbf{S}(p_1^n, p_2^m)$ will be $1$. Otherwise, the far $p_2^m$ is from $l_2$, the closer the score would be $0$. We further add a soft factor $\gamma$ to control the sharpness, thus the  epipolar field is $\mathbf{S}'(p_1^n, p_2^m) = (\mathbf{S}(p_1^n, p_2^m))^\gamma$. 
Figure \ref{fig:epipolar_example} gives a visualization of the epipolar field. Comparing with the epipolar line, the epipolar field model relationships with all pixels in the reference view. We can also reduce it to the epipolar line with a very large $\gamma$. Thus, epipolar field can be considered as a more general form of the epipolar line.  

We then use the epipolar field to guide the learning of $\mathbf{W}_e$ to help the $\mathbf{E_G}_{i}^{n}$ encode correspondence between two views. 
In detail, we let the dot product of $\mathbf{E_G}_{1}^{n}$ and $\mathbf{E_G}_{2}^{m}$ match $\mathbf{S}'(p_1^n, p_2^m)$ with the mean square error loss during the training process: 
\begin{equation}
    \label{eq:loss_pos}
    L_{\text{pos}} = \frac{1}{L^2 } \sum_n^L \sum_m^{L} (\mathbf{E_G}_{1}^{n}  {(\mathbf{E_G}_{2}^{m})}^T -  \mathbf{S}'(p_1^n, p_2^m)  )^2
\end{equation}
Therefore, a high attention score will be achieved along the epipolar line when calculating the cross-view attention maps between $\mathbf{X}_1^{'} + \mathbf{E_{2D}} + \mathbf{E_{G}}_1 $ and $\mathbf{X}_2^{'} + \mathbf{E_{2D}} + \mathbf{E_{G}}_2$, which makes the transformer easy to attend corresponding regions. Moreover, with this soft design, semantic information from offline pixels are still kept, rather than discarded like \cite{he2020epipolar}.

\vspace{-0.5em}
\subsection{Implementation Details}
\vspace{-0.5em}
\paragraph{CNN backbone}
We follow \cite{yang2020transpose} and apply a very shallow CNN architecture as the CNN backbone $\mathcal{F}(\cdot)$, which is the initial part of the ResNet-50 \cite{he2016deep}. Specifically, the number of parameters of the shallow CNN is $1.4$ M, which is just $5.5\%$ of the original Simple Baseline with ResNet-50  ($25.6$M).
The output feature map has size $H=H_I / 8$, $W=W_I / 8$. Thus, the fine-grained local feature information can still be kept.  

\vspace{-1.0em}
\paragraph{TransFusion} 
Following \cite{carion2020end, yang2020transpose}, we set the dimension of the feature embedding $d$ to $256$, the number of heads to $8$, the number of encoder layers $N$ to $3$. Due to the limitation of resource, we only consider the fusion of $2$ neighborhood views, although out framework can be easily extended to more than $2$ views. 

\vspace{-1.0em}
\paragraph{Prediction head}
Given $\tilde{\mathbf{X}_i}$, we first reshape it back to $\tilde{\mathbf{X}_i^{'}} \in \mathbb{R}^{d \times H \times W}$. The prediction head $\mathcal{H}(\cdot)$ applies one deconvolution layer and one $1 \times 1$ convolution layer to predict the heatmap of keypoints. By default, the height and width of heatmaps $\mathbf{H}_i$ are $H_h= H_I / 4$ and $W_h = W_I / 4$.

\vspace{-1.0em}
\paragraph{Loss function} 
The groundtruth heatmap $\mathbf{H}_i \in \mathbb{R}^{k \times H_h \times W_h} $ of 2D keypoints is defined as a 2D a Gaussian centering around each keypoint \cite{wei2016convolutional}. We apply the Mean Square Error (MSE) loss to calculate the difference between the output heatmaps $ \bar{\mathbf{H}}_i$ and $\mathbf{H}_i$. By combining the Equation \ref{eq:loss_pos}, we train the network end-to-end with loss function $L = \frac{1}{HW} \parallel \bar{\mathbf{H}}_i - \mathbf{H}_i \parallel_F^2 + L_{\text{pos}}$.

\vspace{-0.5em}
\section{Experiments}
\vspace{-0.8em}
\subsection{Experimental Settings}
\vspace{-0.5em}
\paragraph{Dataset} 
We conduct extensive experiments on two public multi-view 3D human pose estimation datasets, Human 3.6M \cite{h36m_pami, IonescuSminchisescu11} and Ski-Pose \cite{sporri2016reasearch, rhodin2018learning, gilgien2013determination, gilgien2014effect, gilgien2015determination, fasel2016three, fasel2017joint}. 
(1) The Human 3.6M contains joint annotations of video frames captured by four calibrated cameras in a room. We adopt the same training and test split as in \cite{qiu2019cross, iskakov2019learnable, he2020epipolar}, where subjects 1, 5, 6, 7, 8 are used for training, and 9, 11 are for testing. 
Note that 3D annotations of some scenes of the ’S9’ are damaged \cite{iskakov2019learnable}, 
we exclude these scenes from the evaluation as in \cite{iskakov2019learnable, he2020epipolar}. 
(2) The Ski-Pose dataset aims to help analyze skiers's giant slalom runs with 6 calibrated cameras. It provides six camera views as well as corresponding 3D pose. 
In detail, 8,481 frames are used for training and $1,716$ are used for testing. 
We resize all images to $256 \times 256$ in all experiments.

\vspace{-0.8em}
\paragraph{Training} 
As the training of transformers requires huge datasets \cite{vaswani2017attention, dosovitskiy2020image}, while the scenes of multi-view pose datasets are quite limited, making it difficult to train the transformer from scratch. 
By convention \cite{iskakov2019learnable, he2020epipolar}, 
we use the MS-COCO \cite{lin2014microsoft} pretrained TransPose \cite{yang2020transpose} to initialize our network and fine tune it on the multi-view human pose datasets. Following the settings in \cite{he2020epipolar}, we apply Adam optimizer \cite{kingma2014adam} and train the model for $20$ epochs. The learning rate is initialized with $0.001$ and decays at $10$-th and $15$-th epoch with ratio $0.1$.

\vspace{-0.8em}
\paragraph{Evaluation metrics}
The performance of 2D pose estimation is evaluated by Joint Detection
Rate (JDR), which measures the percentage of the successfully detected keypoints. A keypoints is detected if the distance between the predicted location and the ground truth is within a predefined threshold. The threshold is set to half of the head size for human pose estimation. 
Given the estimated 2D joints of each view, following \cite{qiu2019cross, he2020epipolar}, direct triangulation is used for estimating the 3D poses with respect to the global coordinates. 
The 3D pose estimation accuracy is measured by Mean Per Joint Position Error (MPJPE) between the groundtruth 3D pose and the estimated 3D pose.

\vspace{-0.7em}
\subsection{Results on Human 3.6M}
\vspace{-0.7em}
We compare with two state-of-the-art methods, the crossview fusion \cite{qiu2019cross} and the epipolar transformers \cite{he2020epipolar}. For fair comparison, we use the SimpleBaseline-ResNet50 pretrained on COCO \cite{xiao2018simple} as initialization and then finetuned with their official codes \cite{qiu2019cross, he2020epipolar}.

\vspace{-0.5em}
\paragraph{Quantitative results } 
The results of both 2D and 3D pose estimation are shown in Table \ref{tab:pose2d}. 
We also shown the number of parameters of each model, the MACs (multiply-add operations).
Besides, we also report the inference time to obtain the 3D pose from 4 views on a single 2080Ti GPU of all multiview methods. 
For both 2D and 3D pose estimation, TransFusion consistently outperforms or achieves comparable performance with epipolar transformers \cite{he2020epipolar} and cross-view fusion \cite{qiu2019cross}. 
Note that JDR is a relative loose metric, with a wider threshold which tolerates small errors, so the improvement on 2D is not very obvious. However, on the 3D metric, which directly computes the distance, our improvement is much more significant. Moreover, as in Table \ref{tab:each_pose}, our method can achieve significant improvement on sophisticated poses sequences such as "Phone" and "Smoke", which usually encounters heave occlusions for certain views.  
This result suggests that fusing features from the entire images of other views, instead of just features along the epipolar line \cite{he2020epipolar}, can bring more benefits.  
Besides, comparing to the single view TransPose \cite{yang2020transpose}, our Transfusion can achieve $4.7$ mm gain on 3D. Thus, the improvement is not only from the TransPose architecture, but from the fusion with other views. 
Moreover, our method is lightweight and efficient. It only requires $2.1\%$ (5M / 235M) of the parameters of cross-view fusion \cite{qiu2019cross}. Benefit from the parallel computing of transformers architectures, it further reduces the inference time, while the operation of sampling along epipolar lines \cite{he2020epipolar} is time-consuming.

\begin{table}[ht!]
\centering
\resizebox{0.83\textwidth}{!}{
\begin{tabular}{l|lcc|c|c}
\toprule
Method &  Params & MACs & Inference Time (s) &  JDR (\%) $\uparrow$ & MPJPE (mm) $\downarrow$ \\
\midrule
Single view - Simple Baseline\cite{xiao2018simple}   & 34M &  51.7G  & - & 98.5  & 30.2 \\
Single view - TransPose \cite{yang2020transpose}     & 5M  &  43.6G  & - & 98.6  & 30.5 \\
\midrule
Crossview Fusion \cite{qiu2019cross}                & 235M  & 55.1G  & 0.048 & \textbf{99.4}  & 27.8  \\
Epipolar Transformer \cite{he2020epipolar}          & 34M   & 51.7G  & 0.086 & 98.6 & 27.1 \\
TransFusion                                       & \textbf{5M } & 50.2G & \textbf{0.032} & \textbf{99.4}  & \textbf{25.8} \\
\bottomrule
\end{tabular}
}
\vspace{-0.5em}
\caption{\footnotesize{2D and 3D pose estimation accuracy comparison on Human3.6M. The metric of 2D pose is JDR (\%), and the metric of 3D pose is MPJPE (mm). All networks are pretrained on COCO \cite{lin2014microsoft} and then finetuned on Human 3.6M \cite{h36m_pami}. All images are resized to $256 \times 256$.}}
\label{tab:pose2d}
\end{table}

\begin{table}[ht]
\resizebox{0.99\textwidth}{!}{
\begin{tabular}{l|ccccccccccccccc}
\toprule
Method & Dir & Disc & Eat & Greet & Phone  & Pose & Purch  & Sit & SitD & Smoke & Photo & Wait & WalkD & Walk & WalkT \\
\midrule
Crossview Fusion\cite{qiu2019cross}             & 24.0                                              & 28.8                                             & 25.6                                            & 24.5                                              & 28.3                                              & 24.4                                             & 26.9                                              & \textbf{30.7}                                             & 34.4                                             & 29.0                                                & 32.6                      & 25.1                                             & 24.3                                              & 30.8                                             & 24.9                                              \\
Epipolar transformers \cite{he2020epipolar} & \textbf{23.2}                                             & 27.1                                             & 23.4                                            & 22.4                                              & 32.4                                              & \textbf{21.4}                                              & \textbf{22.6}                                               & 37.3                                            & 35.4                                             & 29.0                                                & 27.7                      & 24.2                                             & \textbf{21.2}                                              & 26.6                                             & \textbf{22.3}                                               \\
TransFusion                  & 24.4                                            & \textbf{26.4}                                    & \textbf{23.4}                                   & \textbf{21.1}                                     & \textbf{25.2}                                     & 23.2                                             & 24.7                                              & 33.8                                   & \textbf{29.8}                                    & \textbf{26.4}                                     & \textbf{26.8}             & \textbf{24.2}                                    & 23.2                                              & \textbf{26.1}                                    & 23.3           \\
\bottomrule
\end{tabular}
}
\vspace{-0.5em}
\caption{\footnotesize{The MPJPE of each pose sequence on Human 3.6M.}}
\label{tab:each_pose}
\end{table}

\begin{figure}[!ht]
    \centering
    \includegraphics[width=0.7\textwidth]{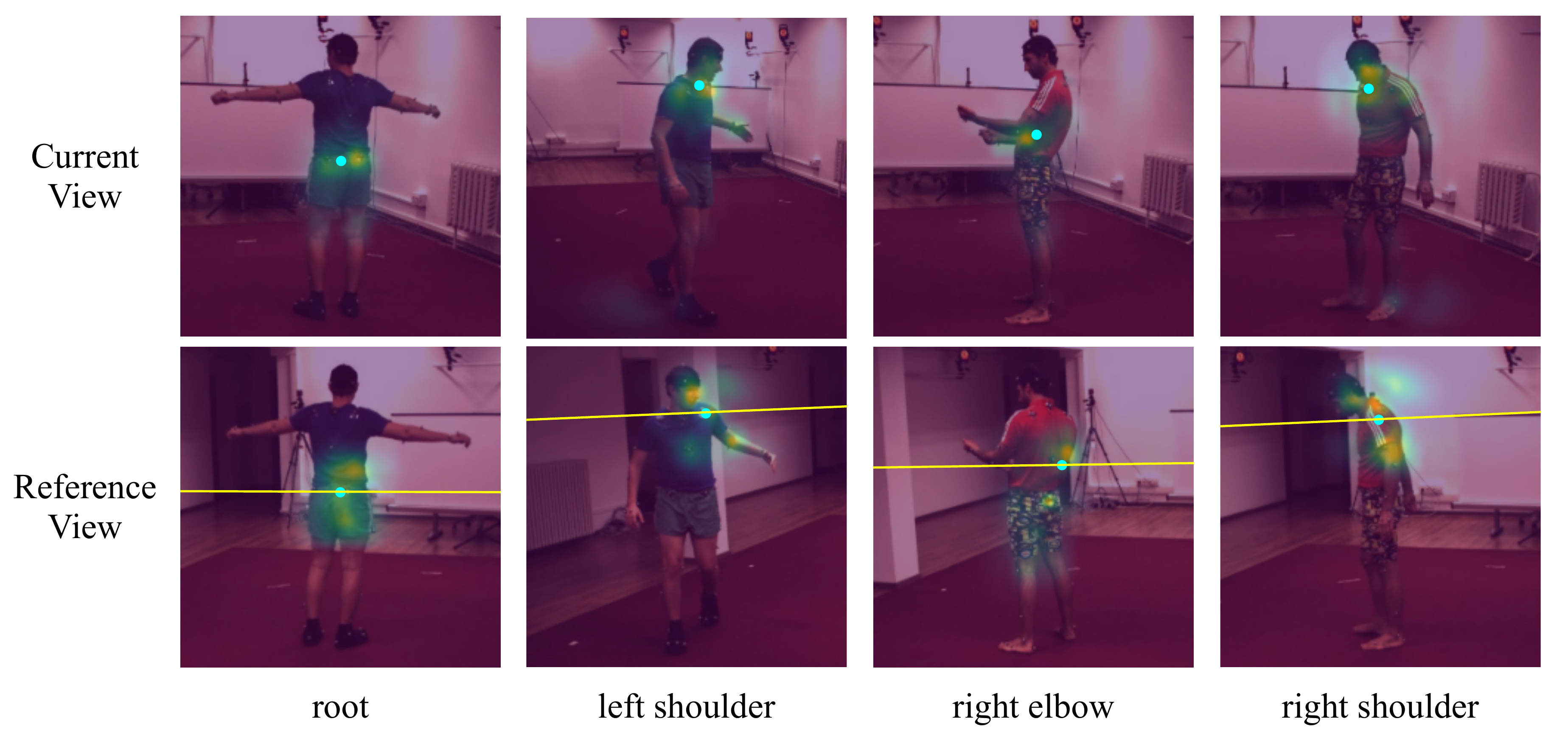}
    \vspace{-1 em}
    \caption{\footnotesize{Visualization of attention maps on Human 3.6M test set. The cyan dots are groundtruth. Given the query pixel, the first rows are attention maps of current views, and the second rows are attention maps of reference views. We also visualize the epipolar line (yellow) for comparison. }}
    \label{fig:more_att_map}
\end{figure}

\vspace{-0.8em}
\paragraph{Visualization of Attention maps}
Given the query pixel in one view, we further visualize the attention maps on both views. We show our results in Figure \ref{fig:more_att_map}.
It is observed that on the view itself, typically the attention map is around the joints. If the query joint is occluded, it may resort to joints on the other side of the symmetry \cite{yang2020transpose}. On the neighboring view, the network usually not just attends the corresponding keypoint, but attends the whole limbs, which cannot be located by the epipolar line. Previous methods based on epipoar line \cite{he2020epipolar} actually miss this important clue.

\begin{figure}[!ht]
    \centering
    \includegraphics[width=0.85\textwidth]{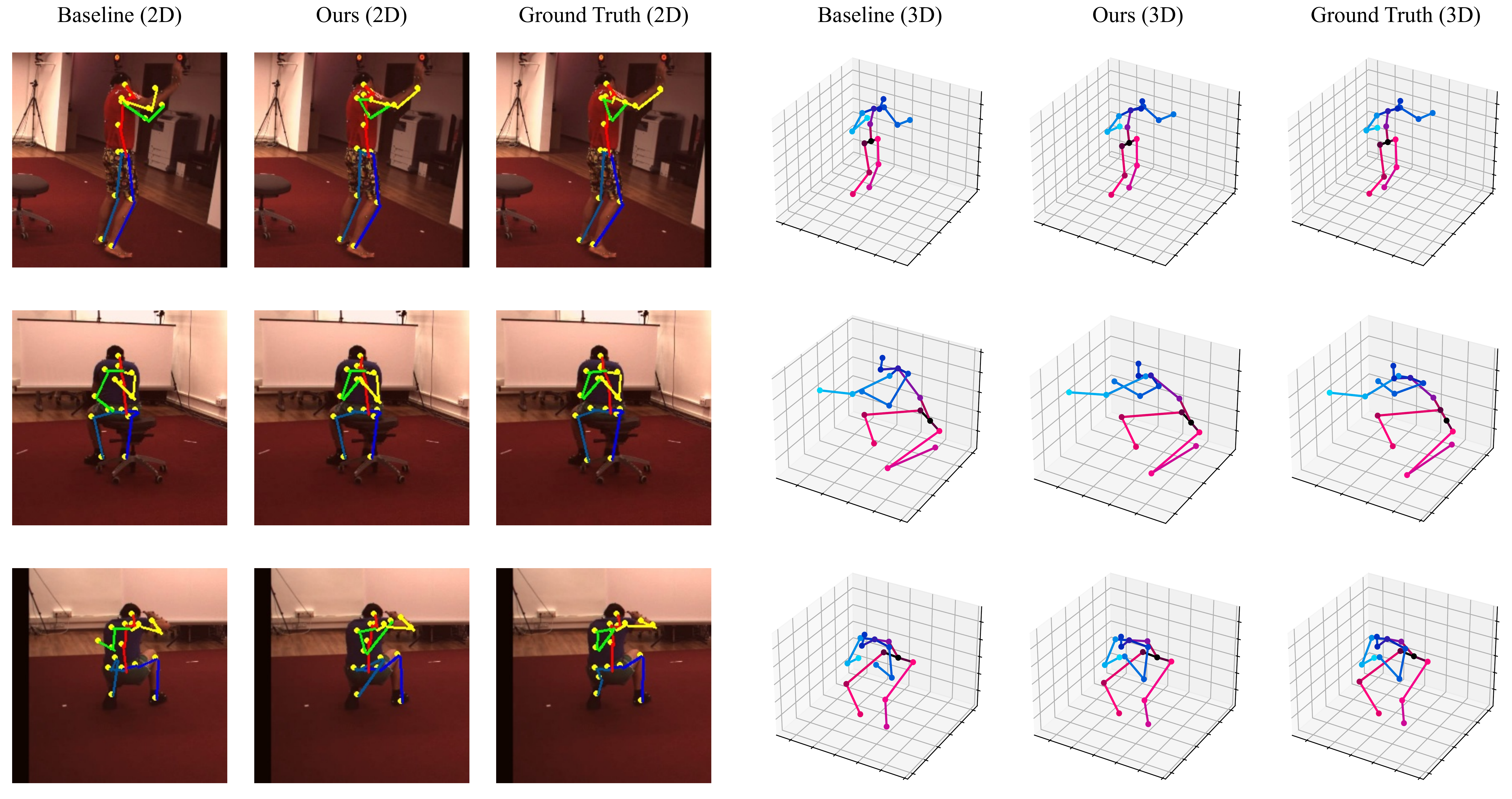}
    \vspace{-1 em}
    \caption{\footnotesize{Visualization of predictions for Human 3.6M. Both skeletons of 2D keypoints on the image and 3D pose in the space are presented. }}
    \label{fig:examples_2d_3d}
\end{figure}

\vspace{-0.8em}
\paragraph{Qualitative results }
We also present examples of predicted 2D keypoints on the image and 3D pose in the space, and compare our methods with baseline methods \cite{qiu2019cross}. As in Figure \ref{fig:examples_2d_3d}, even if the entire arms (green line) are occluded, our method still predicts the 2D keypoints correctly by fusing information from the reference view, and further gives a better 3D pose.

\vspace{-0.5em}
\subsection{Ablation Studies}
\vspace{-0.5em}
\paragraph{Geometry Positional Encoding}
We conduct ablation studies on the GPE to verify its significance. In detail, we consider 3  settings: 1) training without 3D geometry positional encoding, 2) applying a learnable 3D positional encoding, \ie directly learn $\mathbf{E_G}_i$ from scratch 3) training the 3D GPE without epipolar field constraints $L_{\text{pos}}$. 
Table \ref{tab:ablation_gpe} presents the results. Without the 3D location information, the performance of 1) and 2) are even worse than the single view TransPose, we hypothesize that the 2D sine PE makes the transformer easy to attend the same pixel location of all views, and the learned 3D PE is easy to overfit the training examples. 
Without $L_{\text{pos}}$, the error will also increase. Thus the guide from the epipolar field is favorable, as it imposes correspondence for cross-view attention. 

\begin{table}[ht!]
\centering
\resizebox{0.79\textwidth}{!}{
\begin{tabular}{l|c|c}
\toprule
Method &  2D Pose / JDR (\%) $\uparrow$ & 3D Pose / MPJPE (mm) $\downarrow$ \\
\midrule
TransFusion - without 3D positional encoding      & 98.5  & 35.9 \\
TransFusion - learnable 3D positional encoding    & 96.0  & 57.3 \\
TransFusion - GPE without $L_{\text{pos}}$            & 99.3  & 26.8  \\
TransFusion                                       & \textbf{99.4}  & \textbf{25.8} \\
\bottomrule
\end{tabular}
}
\vspace{-1 em}
\caption{\footnotesize{ Ablation studies on different types of 3D positional encoding } }
\label{tab:ablation_gpe}
\end{table}

\begin{figure}[h]
\centering  
\subfigure[{ \footnotesize{Soft Factor}}]{
\label{fig:gamma}
\includegraphics[width=0.3\textwidth]{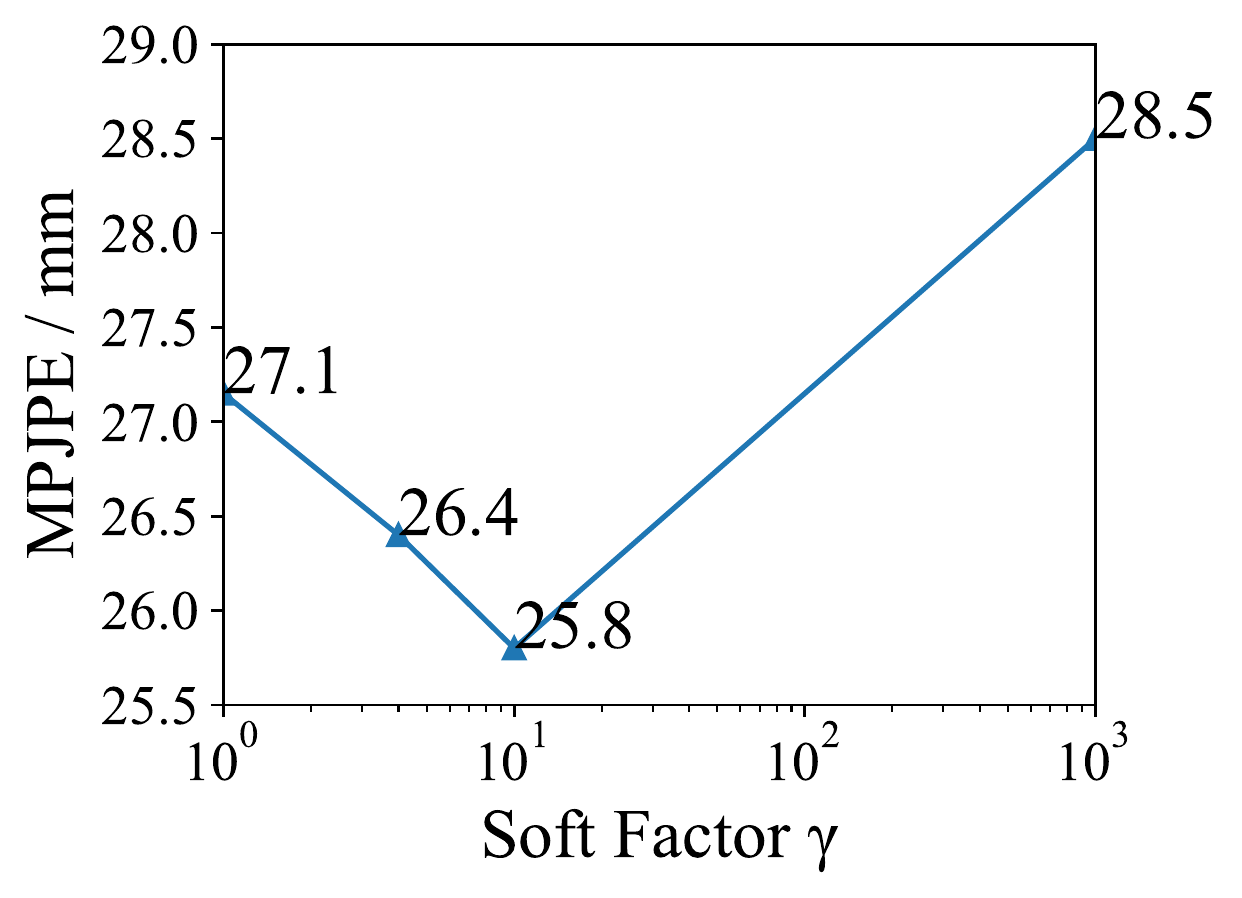}}
\subfigure[{ \footnotesize{Number of encoder layers}}]{
\label{fig:num_layers}
\includegraphics[width=0.3\textwidth]{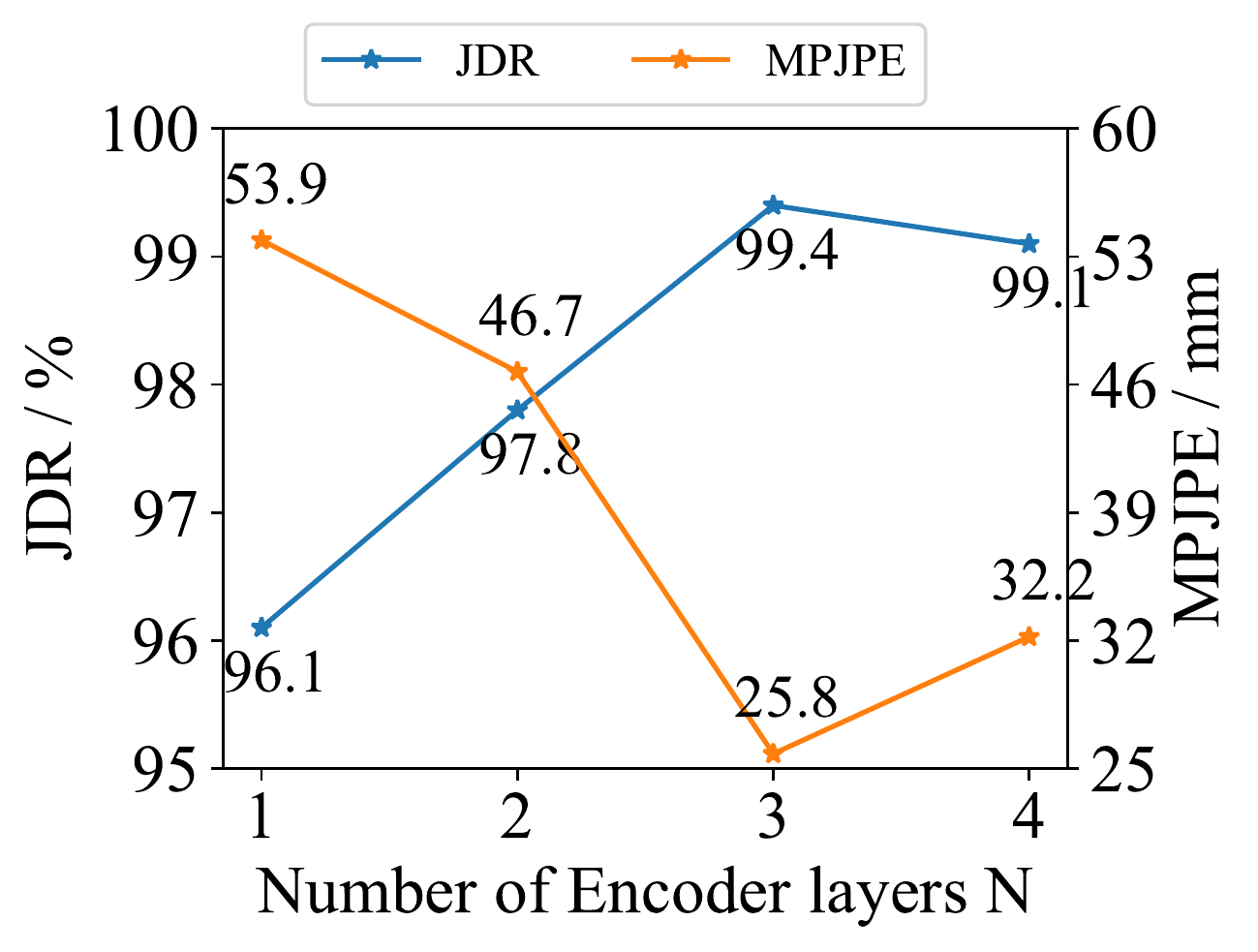}}
\subfigure[{ \footnotesize{Number of heads}}]{
\label{fig:num_heads}
\includegraphics[width=0.3\textwidth]{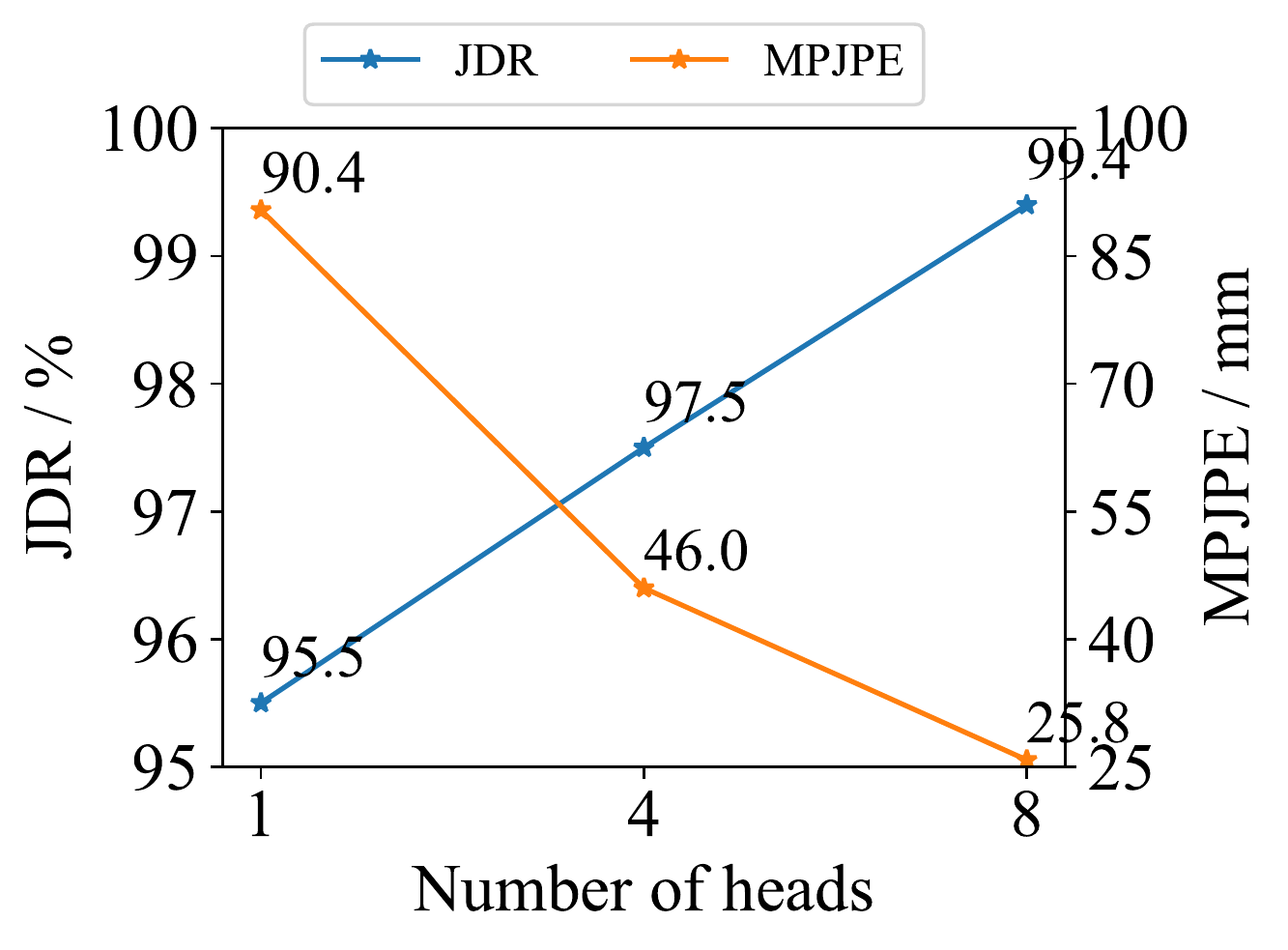}}
\vspace{-1.0em}
\caption{ \small{Ablation studies on the soft factor $\gamma$ of the epipolar field, the number of transformer encoder layers $N$, and the number of transformer heads.}}
\vspace{-0.5em}
\end{figure}

\vspace{-1.0 em}
\paragraph{Soft Factor $\gamma$}
We also try different values of the soft vector $\gamma$, results are shown in Figure \ref{fig:gamma}. With a small $\gamma$, the epipolar field assign all locations with relative high probabilities, the performance are slightly worse (1.3 mm drop).  
While with a huge $\gamma=1000$, the epipolar field reduces to the hard epipolar line, and the performance drops $2.7$ mm. Thus, we verify the effectiveness of our epipolar field compared with hard-coded epipolar line.

\vspace{-1.0 em}
\paragraph{Transformer architecture} We study how performance scales with the size of the transformer. As in Figure \ref{fig:num_layers}, with the number of layers $N$ increasing, the performance improves significantly, as the learning ability of transformer is more powerful with more parameters. 
But when $N>3$, it tends to saturate or degenerate. We hypothesize that the transformer is easy to overfit when the size is too huge. 
Meanwhile, as in Figure \ref{fig:num_heads},  with the number of heads increases, the performance also improves gradually, as more heads can help attend different features \cite{vaswani2017attention}.  
In summary, our choice with $N=3$ and 8 heas are reasonable.

\begin{table}[ht!]
\centering
\resizebox{0.75\textwidth}{!}{
\begin{tabular}{l|c|c}
\toprule
Method  & 2D Pose / JDR (\%) $\uparrow$ & 3D Pose / MPJPE (mm) $\downarrow$ \\
\midrule
Single view - Simple Baseline \cite{xiao2018simple}  & 94.5  & 39.6 \\
Epipolar Transformer \cite{he2020epipolar}           & 94.9  & 34.2  \\
TransFusion                                          & \textbf{96.0} & \textbf{31.6}  \\
\bottomrule
\end{tabular}
}
\vspace{-1em}
\caption{\footnotesize{2D and 3D pose estimation accuracy comparison on Ski-Pose.} }
\label{tab:skipose}
\end{table}

\vspace{-1.0em}
\subsection{Results on Ski-Pose Dataset}
\vspace{-0.7em}
We further apply TransFusion on the Ski-Pose dataset to verify its generalization ability. Results are presented in Table \ref{tab:skipose}. 
In the settings with six cameras, the Crossview Fusion is too huge (537M) to train on the 2080Ti GPU. Similar to Human 3.6M, TransFusion still outperform or achieve comparable performance with other fusion methods, while it is much lightweight.
Thus, our method is also effective in outdoor multi-view settings.

\vspace{-1.0em}
\section{Conclusion}
\vspace{-0.5em}
In this paper, we apply the transformer to the multi-view 3D human pose estimation for the first time. Inspired by multi-modal transformers, we propose the TransFusion network, a lightweight architecture to integrate cues from both self views and reference views. Furthermore, we propose the epipolar field, and apply it to the 3D  positional encoding to encode correspondence between two views explicitly. Experimental results shows that our method outperform previous fusion methods but with a more light weighted network. 
In the future we plan to apply our TransFusion to regress the 3D locations with multi-view inputs in an end-to-end way to further improve 3D predictions.

\bibliography{egbib}

\end{document}